\documentclass[runningheads]{llncs}
\usepackage{graphicx}
\usepackage{cite}
\usepackage{hyperref}
\usepackage[misc]{ifsym}
\usepackage{bbm}
\usepackage{url}
\usepackage{color}

\makeatletter
\newcommand{\printfnsymbol}[1]{%
  \textsuperscript{\@fnsymbol{#1}}%
}

\makeatother
\begin{document}
\title{Enhancing Data Diversity for Self-training Based Unsupervised Cross-modality Vestibular Schwannoma and Cochlea Segmentation}

\titlerunning{Enhancing Data Diversity for Self-training Cross-modality Segmentation}


\author{Han Liu\inst{1}\thanks{equal contribution}\Letter
\and Yubo Fan\inst{1}\printfnsymbol{1} 
\and Ipek Oguz\inst{1,2}
\and Benoit M. Dawant\inst{2}}

\authorrunning{H. Liu et al.}

%
\institute{Department of Computer Science, Vanderbilt University\and 
Department of Electrical and Computer Engineering, Vanderbilt University\\
\email{han.liu@vanderbilt.edu}}

\maketitle              
\begin{abstract}
Automatic segmentation of vestibular schwannoma (VS) and cochlea from magnetic resonance imaging can facilitate VS treatment planning. Unsupervised segmentation methods have shown promising results without requiring the time-consuming and laborious manual labeling process. In this paper, we present an approach for VS and cochlea segmentation in an unsupervised domain adaptation setting. Specifically, we first develop a cross-site cross-modality unpaired image translation strategy to enrich the diversity of the synthesized data. Then, we devise a rule-based offline augmentation technique to further minimize the domain gap. Lastly, we adopt a self-configuring segmentation framework empowered by self-training to obtain the final results. On the CrossMoDA 2022 validation leaderboard, our method has achieved competitive VS and cochlea segmentation performance with mean Dice scores of 0.8178 ± 0.0803 and 0.8433 ± 0.0293, respectively.


\end{abstract}

\section{Introduction}
Vestibular schwannoma (VS) is a benign tumor that stems from an overproduction of Schwann cells. It develops from the vestibular nerve which connects the brain and the inner ear and its common symptoms include hearing loss, dizziness, and tinnitus \cite{VS_web}. Magnetic resonance imaging (MRI) is crucial for diagnosis and surveillance of VS and contrast enhanced T1 weighted (ceT1) MRI is currently the most commonly used protocol. However, this involves gadolinium, which may produce side effects ranging from mild to severe. As a possible noncontrast and lower-cost alternative, high-resolution T2-weighted (hrT2) imaging has shown promises for follow-up surveillance scans \cite{coelho2018mri, shapey2019artificial, pizzini2020usefulness}.

To facilitate the clinical workflow, automatic methods to segment the VS in ceT1 and hrT2 have recently emerged \cite{shapey2019artificial, wang2019automatic,zhu2022acoustic}. However, training supervised VS segmentation models requires manual annotation, which is expensive and time-consuming. Weakly-supervised \cite{dorent2020scribble} and unsupervised VS segmentation methods \cite{dorent2021crossmoda} have thus drawn increasing interest in the community. In the CrossMoDA 2021 challenge \cite{dorent2021crossmoda}, participants were given the task of unsupervised cross-modality VS and cochlea segmentation, i.e.,  segmenting these two structures in hrT2, but annotations were only provided in unpaired ceT1 images in the training set. In the CrossMoDA 2022 challenge, an additional set of ceT1 and hrT2 images are acquired at another MRI site. Therefore, two types of domain gaps exist in this challenge due to the difference in (1) acquisition sites, i.e., site A vs.\ site B, and (2) MRI modalities, i.e., ceT1 vs.\ hrT2, as shown in Figure 1. Both domain gaps need to be addressed to achieve robust segmentation performance in CrossMoDA 2022.

In this paper, we present our solution to the segmentation task of CrossMoDA 2022. We approach this task as an unsupervised domain adaptation (UDA) problem where we first train cross-site unpaired image translation models to generate pseudo target domain (hrT2) images, then apply a rule-based augmentation to the pseudo hrT2 images, and finally train nnU-Net \cite{isensee2021nnu} segmentation models using a self-training scheme. The results on the challenge leaderboard showed that our method has achieved promising segmentation performances on both VS and cochlea.

\begin{figure}[t]
\includegraphics[width=1\columnwidth]{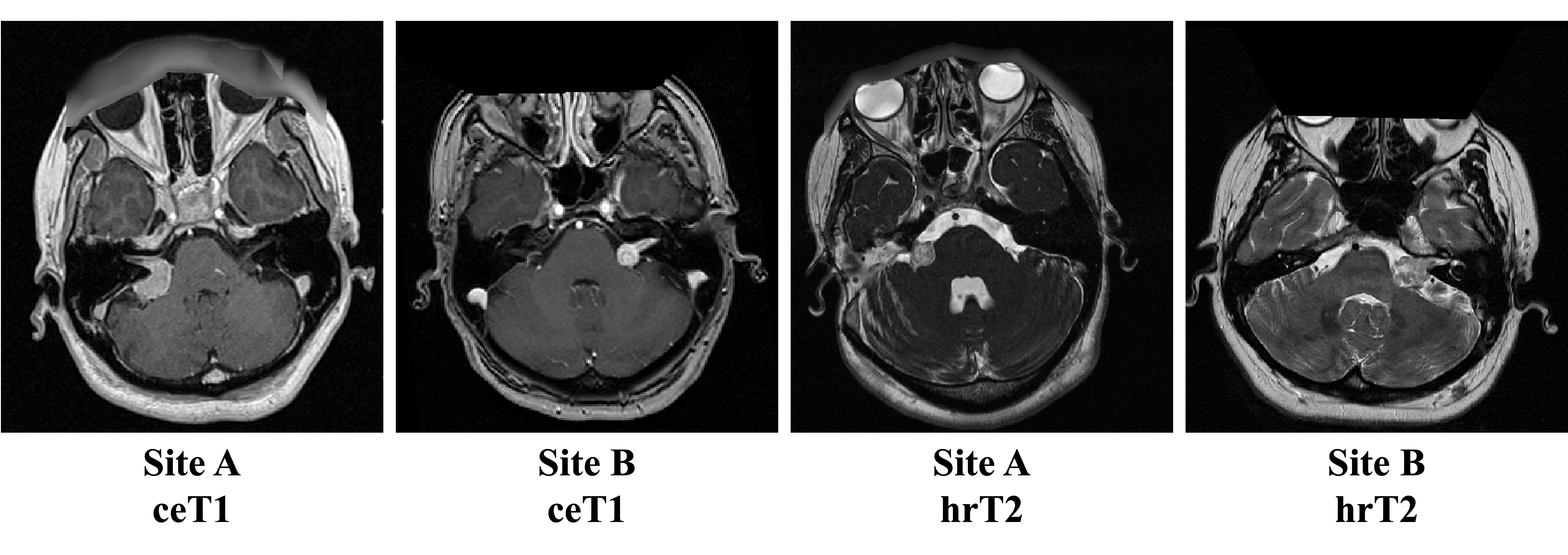}
\centering
\caption{Two types of domain gaps exist in CrossMoDA 2022 due to the difference in (1) acquisition sites and (2) MRI modalities.} 
\label{fig1}
\end{figure} 

\section{Related Work}
\subsection{Unsupervised Domain Adaptation (UDA)}
The performance of machine learning models can be affected by data distribution shift between the training/source dataset and test/target dataset \cite{Guan_Liu_2022}. UDA refers to the task of improving model performance on target domain data when their label is not available. Feature alignment-based methods aim to learn domain-invariant features across different domains. Domain Adversarial Neural Network (DANN) \cite{Ganin_Ustinova_Ajakan_Germain_Larochelle_Laviolette_Marchand_Lempitsky_2017} is a representative architecture that utilizes adversarial learning between a feature extractor and a domain discriminator. The domain discriminator learns to differentiate whether the extracted features are from the source or target domain, and the feature extractor learns domain-invariant features such that it can fool the discriminator. Another way to align features is by optimizing some divergence metrics \cite{Long_Cao_Wang_Jordan_2015,Yan_Ding_Li_Wang_Xu_Zuo_2017} between the source and target domains. With the emergence of unpaired image-to-image translation approaches such as CycleGAN \cite{zhu2017unpaired} and UNIT \cite{Liu_Breuel_Kautz_2017}, image-level alignment is also used to tackle the UDA problem \cite{Hoffman_Tzeng_Park_Zhu_Isola_Saenko_Efros_Darrell_2018}. By utilizing a cycle-consistency loss between the input and the reconstructed images, the models are trained without paired data. Then, image-level alignment can be achieved by generating pseudo target domain images from the source domain images. In this work, we adopt image-level alignment paradigm to bridge the domain gap between ceT1 and hrT2.

\subsection{Self-training in UDA}
Self-training strategies have shown promising results in the field of UDA by fully utilizing the unlabeled target domain data. It has been shown to improve the performance of the segmentation models by fine-tuning them on the target images with pseudo labels. Zou et al.\ \cite{zou2018unsupervised} propose a confidence regularized self-training framework which encourages the smoothness of the network output and reduces the confidence in false positives during training. Yu et al.\ \cite{Yu_Zhang_Dong_Hu_Dong_Zhang_2021} use a portion of the pseudo labels with high probability to iteratively fine-tune the model and achieve superior performance on a UDA dataset. This strategy is also used by the top teams in CrossMoDA 2021 \cite{dorent2021crossmoda}, further demonstrating its effectiveness in the UDA problem.

\section{Methods}

\subsection{Overview}
In this study, we propose to tackle the UDA segmentation problem by following the popular ``synthesis-then-segmentation" training strategy \cite{dong2021unsupervised,shin2022cosmos,choi2021using,lastyear}. Specifically, we perform unpaired image translation to synthesize the pseudo target domain images in a cross-site and cross-modality fashion. Then we devise a rule-based offline augmentation method to further increase the data variability. Lastly, we adopt the nnU-Net framework
and perform self-training to train a target-domain segmentation model with both the synthetic and real target domain data.

\begin{figure}[t]
\includegraphics[width=1\columnwidth]{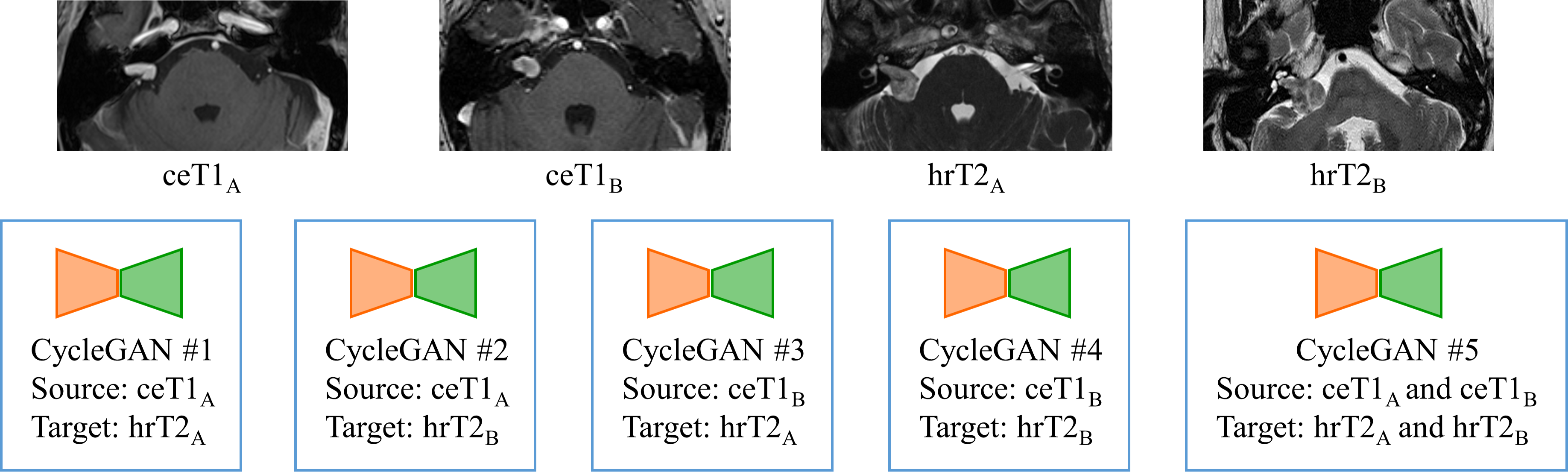}
\centering
\caption{The diagram of cross-site cross-modality unpaired image translation.} \label{fig2}
\end{figure} 

\subsection{Cross-site Cross-modality Unpaired Image Translation}
Since our task is to segment the VS and cochlea in hrT2 images while their labels are only available in the unpaired ceT1 images, we adopt an unpaired image translation method, i.e., CycleGAN \cite{zhu2017unpaired}, to bridge the gap between the two modalities and synthesize pseudo hrT2 from ceT1. Then the pseudo hrT2 images with the annotations from the corresponding ceT1 can be utilized to train segmentation models. However, MRIs (both ceT1 and hrT2) in the provided dataset were acquired from two different sites, and we observe slight appearance differences within each modality (especially hrT2) across sites. To enrich the diversity of the pseudo hrT2 images, we synthesize them using ceT1 within and across sites. Specifically, as shown in Figure \ref{fig2}, five CycleGAN models are trained with different source/target domain data configurations to generate pseudo hrT2 images. We use subscripts to denote images from site A or site B, e.g., hrT2\textsubscript{A} means hrT2 from site A. Compared to only generating within-site pseudo images, i.e., pseudo hrT2\textsubscript{A} from ceT1\textsubscript{A} (CycleGAN \#1) and pseudo hrT2\textsubscript{B} from ceT1\textsubscript{B} (CycleGAN \#4), our cross-site training scheme can generate twice more pseudo hrT2 data for the downstream task of nnU-Net segmentation. Note that CycleGAN \#5 has the same network architecture as the other four, and the only difference is that its training set contains images from both site A and site B.

\subsection{Rule-based Offline Augmentation for VS and Cochlea}
\begin{figure}[t]
\includegraphics[width=1\columnwidth]{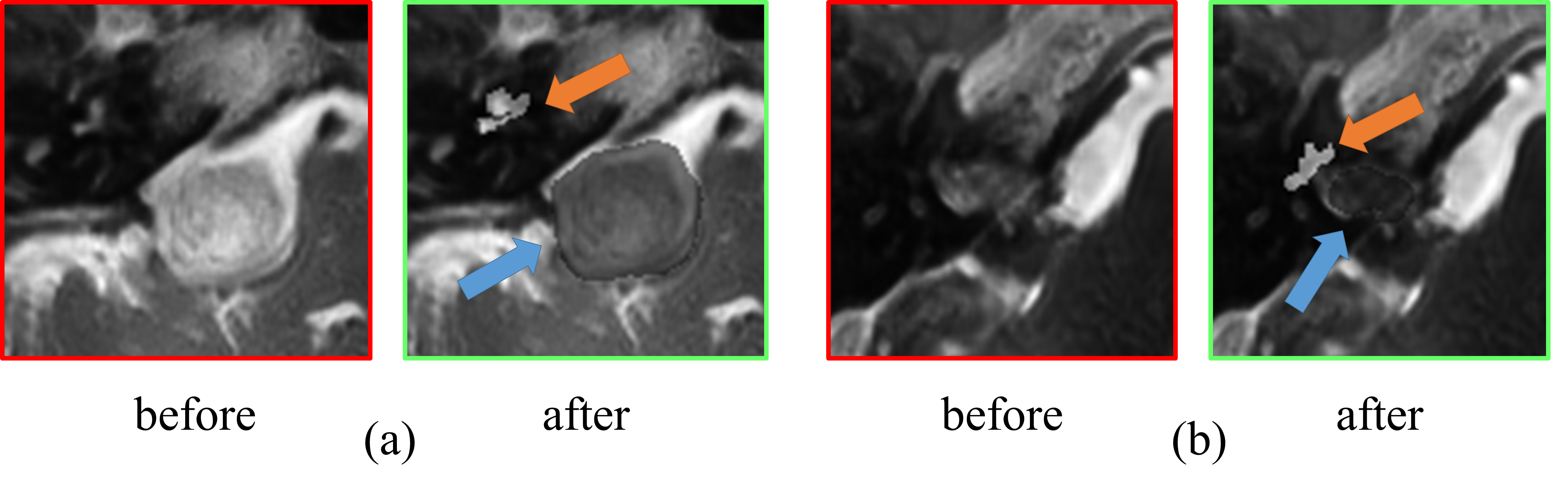}
\centering
\caption{An illustration of the rule-based offline augmentation technique on two cases. Orange and blue arrows represent the augmented cochlea and VS, respectively.} 
\label{fig3}
\end{figure}

Thanks to the unpaired image translation, the domain gap between ceT1 and hrT2 can be substantially reduced at the image-level by generating pseudo hrT2 images. However, there is still a domain gap between the pseudo and real hrT2, as we observe that the cochlea and VS in the pseudo hrT2 images do not have the same intensity characteristics as in the real hrT2 images. To overcome this issue, we propose to adjust the intensities of the VS and cochlea regions in pseudo hrT2 images to further minimize the domain gap, as shown in Figure \ref{fig3}. As suggested by \cite{choi2021using}, for VS, we reduce the signal intensity of the voxels that are labeled as VS by 50\% to further increase the heterogeneity of VS signals \cite{choi2021using}. This VS augmentation is randomly applied to 50\% of our pseudo hrT2 images. Cochleae typically have high signal intensities in hrT2 images, with values that were empirically found to be within the  [85th, 95th] intensity percentile. However, we observe that the generated cochlea in pseudo hrT2 images can have low intensities or can even be absent. To improve the appearance of pseudo hrT2 images with a mean cochlea intensity lower than the 85th intensity percentile, we replace the original intensity of the cochlea voxels by a value randomly sampled from a uniform distribution, which is bounded by the 85th and 95th intensity percentile of that image. Lastly, the augmented cochlea region is smoothed by a 3D Gaussian kernel to further refine the cochlea appearance. The cochlea augmentation is applied to all the pseudo hrT2 images.

\subsection{Segmentation with nnU-Net and Self-training}
With the augmented pseudo hrT2 images, we can train a segmentation model supervised by the paired ceT1 labels. Here, we utilize a popular self-configuring segmentation framework, i.e., nnU-Net, for supervised learning. However, with this approach the real hrT2 images are not involved in training the segmentation model due to the lack of labels. To tackle this problem, we adopt a self-training strategy to make use of the unlabeled real hrT2 images. Specifically, we firstly use the model trained on pseudo hrT2 images to obtain the pseudo labels of the unlabeled real hrT2 images. Next, we re-train the nnU-Net with the combined data, i.e., pseudo hrT2 images with real ceT1 labels and real hrT2 images with pseudo labels. Note that we perform self-training several (in our case, 3) times, since we observe improvements on the validation leaderboard each time the network is re-trained with the updated pseudo labels produced by self-training.

\section{Experiments and Results}
\subsection{Dataset}
The dataset was released by the MICCAI challenge CrossMoDA 2022 \cite{shapey2021segmentation}. 105 ceT1 and 105 hrT2 unpaired images were obtained on a 32-channel Siemens Avanto 1.5T scanner, which are called ``London data" or site A data. Another 105 ceT1 and 105 hrT2 unpaired images were obtained on a Philips Ingenia 1.5T scanner, which are called ``Tilburg data" or site B data. Image resolutions are different for images from different sequences and sites. The manually segmented mask of the VS and cochlea for the 210 ceT1 images are provided.

\subsection{Pre-processing and Network Training}
We observe that the fields of view and resolutions vary substantially across different sites and modalities. Inspired by \cite{lastyear}, to discard the irrelevant brain regions in our task, we first crop each MRI scan into a region of interest (ROI) by rigid registration with an atlas. Specifically, we use 4 atlases (2 sites × 2 modalities) and we register each scan with the atlas from the same site and modality. Moreover, to avoid losing information from high-resolution images, we resample all the images and masks to the highest resolution of the images in the challenge dataset, i.e., [0.4102, 0.4102, 1.0] mm. 

For CycleGANs, we train the 5 models (as in Figure \ref{fig2}) in both 2D and 3D settings. The patch size for 2D CycleGANs is 256 $\times$ 256 and for 3D is 112 $\times$ 112 $\times$ 24. Random cropping or zero padding is used when the dimension of the resampled ROI does not match the patch size during training. We use Adam optimizer \cite{Kingma_Ba_2017} with a fixed learning rate of 2e-4. The training is stopped after 100 epochs for the 2D CycleGANs and after 1000 epochs for the 3D CycleGANs. Sliding window inference with an overlapping ratio of 0.8 is used to generate the final synthesis results. Synthesized 2D slices are then merged into the 3D volumes based on their original positions.

For nnU-Net training, we use the network architecture and patch size provided by the 3D fullres mode. Five-fold cross-validation is used. The Dice loss + cross-entropy loss is used as the loss function. The SGD optimizer with an initial learning rate of 1e-2 is used and the learning rate is decayed by a polynomial function. We do not apply connected component analysis to post-process the segmentation results, as we find that the impact of this post-processing operation on the segmentation results varies across folds.

Overall, we use 4 training stages. In stage 1, we train the nnU-Net with the pseudo hrT2 images generated from 2D and 3D CycleGANs. In stage 2, we replace the images from 2D CycleGANs with real hrT2 images for self-training. In stage 3, we apply offline VS augmentation and oversample the pseudo hrT2 images with small tumors, as we observe on the validation leaderboard that the small-tumor segmentation performance of our network is low. We oversample the pseudo hrT2 images with small tumors based on the VS labels from ceT1 till the training data size reaches 1000 (maximum number that nnU-Net allows). In stage 4, we perform another round of self-training and use the ensemble model from stages 2, 3, and 4 as our final model by averaging softmax probabilities. 

\begin{figure}[t]
\includegraphics[width=0.95\columnwidth]{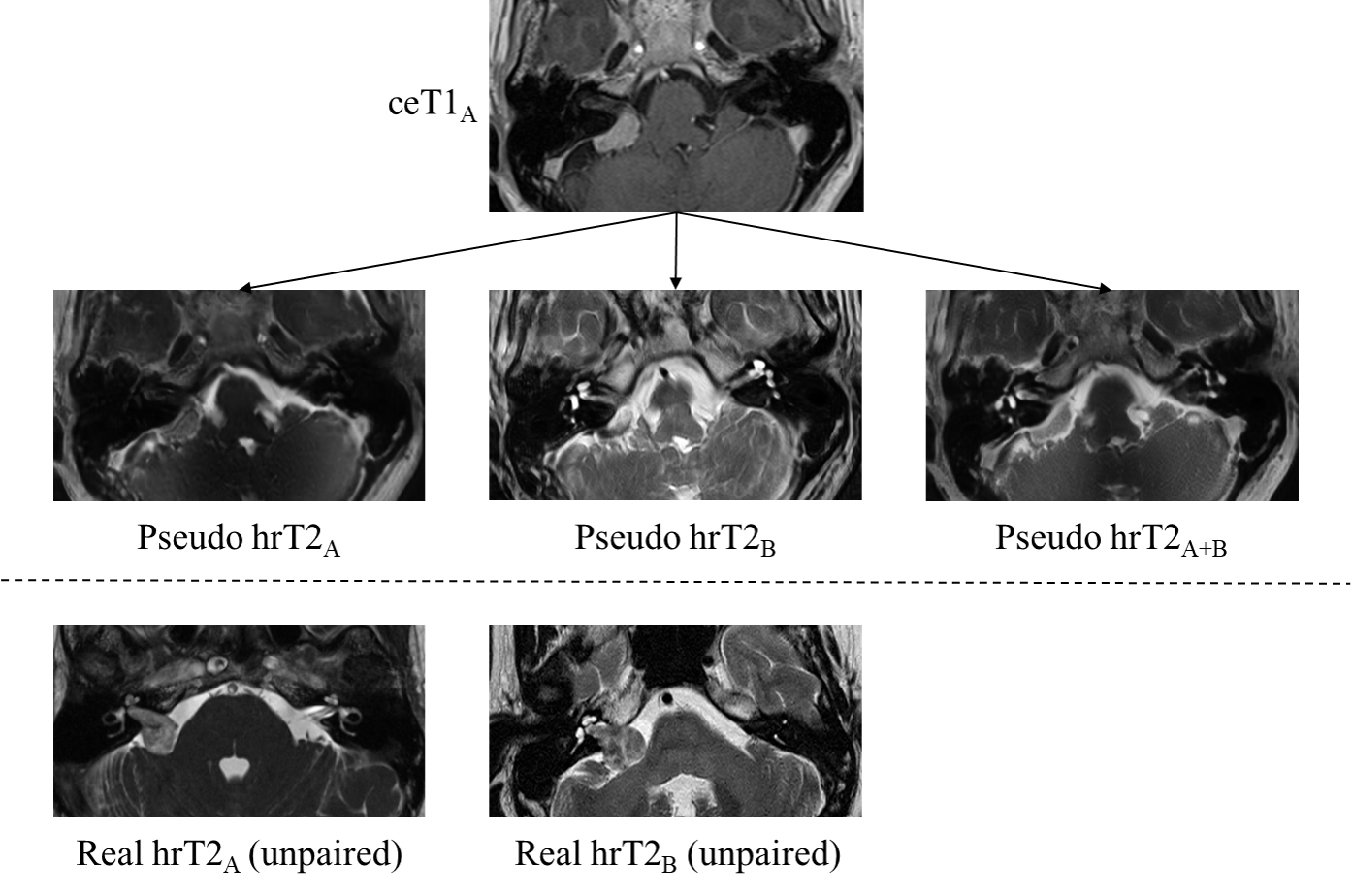}
\centering
\caption{Synthesis results of the cross-site cross-modality unpaired image translation. First row: source domain (ceT1) image. Second row: pseudo target domain (hrT2) images. Third row: real target domain (hrT2) images as reference.} \label{fig4}
\end{figure} 

\subsection{Experimental Results}
\subsubsection{Synthesis Results.}
In Figure \ref{fig4}, we qualitatively show the effectiveness of the cross-site cross-modality unpaired image translation. We select a representative ceT1 image from site A and generate the pseudo hrT2 images using CycleGAN \#1, \#2, and \#5, which generate pseudo hrT2 from site A, site B, and a combination of site A and site B, respectively. Two representative real hrT2 images from site A and from site B are provided in Figure \ref{fig4} for comparison. We observe the consistency of the overall image contrast characteristics between the pseudo and real images from the same site.

\subsubsection{Segmentation Results.} 
\begin{figure}[t]
\includegraphics[width=1\columnwidth]{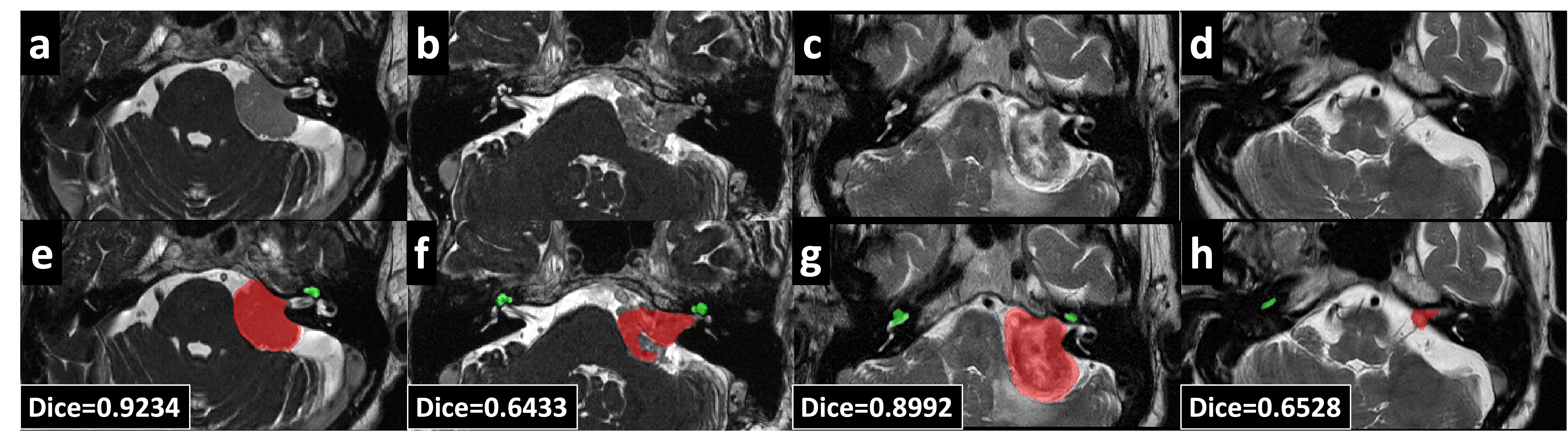}
\centering
\caption{Qualitative results on the validation set. The first row and the second row correspond to the hrT2 images and their segmentation results, respectively. Note that a, b and c, d are from different sites.} \label{fig5}
\end{figure}

For quantitative evaluation, we submit our segmentation results to the validation leaderboard, where the Dice score and average symmetric surface distance (ASSD) between segmentation results and the ground truth are computed. In Table 1, we show the segmentation results for both VS and cochlea obtained at each training stage (the configurations are described in detail in Section 3.2). We notice that the largest improvement is observed from stage 1 to stage 2, when the model is trained on real target domain images for the first time. Moreover, the effectiveness of our proposed offline VS augmentation and small-tumor oversampling is demonstrated by the improvement observed from stage 2 to stage 3. Lastly, the mean Dice scores and ASSDs achieved by our final ensemble model are 0.8178 ± 0.0803 and 0.8433 ± 0.0293, and 0.6673 ± 0.2713 mm and 0.2053 ± 0.1489 mm for VS and cochlea, respectively. For qualitative evaluation, we visualize the segmentation results in Figure \ref{fig5} for four representative cases (a and b are from site A; c and d are from site B). We posit that the poor VS segmentation, i.e., f and h, is due to the uncommon appearances of VS in the validation set, which can be difficult to synthesize if such appearances rarely exist in our target domain training dataset.

\begin{table}[ht]
\centering
\caption{Quantitative results on the validation leaderboard}
\begin{tabular}{cccccc}

\hline
  & & \multicolumn{2}{c}{Dice$\uparrow$ (\%)} & \multicolumn{2}{c}{ASSD$\downarrow$ (mm)}  \\ \cline{3-6} 
  & Stage  & VS              & Cochlea             & VS                 & Cochlea            \\ \hline
  & 1   & 73.54±24.26        & 81.91±4.33    & 1.89±8.41          & 0.25±0.15          \\
  & 2   & 77.97±18.29        & 82.42±4.01    & 0.72±0.29          & 0.23±0.15          \\
  & 3   & 80.37±9.64         & \textbf{84.46±2.97}    & 1.14±1.73          & 0.22±0.18          \\
  & ensemble   & \textbf{81.78±8.03}         & 84.33±2.93    & \textbf{0.67±0.27}          & \textbf{0.21±0.15}  \\ \hline
\end{tabular}
\end{table}

\section{Discussion and Conclusion}
In the CrossMoDA 2021 challenge, we have observed that there were three important components used by the top teams including (1) segmentation via nnU-Net \cite{choi2021using,dong2021unsupervised,shin2022cosmos}, (2) self-training with real target domain data \cite{shin2022cosmos,dong2021unsupervised}, and (3) data augmentation for VS and cochlea \cite{choi2021using,li2022unsupervised}. Based on this observation, our proposed method for CrossMoDA 2022 also incorporates these components to achieve competitive performance on the leaderboard. To address the additional domain gap (multi-site MRI) in CrossMoDA 2022, we synthesize site-specific pseudo target domain images with multiple CycleGAN models. Nevertheless, we conjecture that the segmentation performance might be further improved if we also train site-specific target domain segmentation models; this will be investigated.

In conclusion, we proposed a solution to tackle the UDA problem for VS and cochlea segmentation in CrossMoDA 2022. We developed a cross-site cross-modality unpaired image translation strategy to enrich the diversity of the synthesized data and a rule-based offline augmentation method to further minimize the domain gap. Lastly, we empowered the nnU-Net by self-training to make use of the unlabeled data. According to the validation leaderboard, our method has achieved a promising segmentation performance on both VS and cochlea.

\section{Acknowledgements}
This work was supported in part by the NIH grants NIDCD-DC014462, NIDCD-DC008408, NIBIB-T32EB021937, and by the Advanced Computing Center for Research and Education (ACCRE) of Vanderbilt University. This publication contents are solely the responsibility of the authors and do not necessarily represent the official views of the founding institutes.

\bibliographystyle{splncs04}
\bibliography{references.bib}

\begin{thebibliography}{10}
\providecommand{\url}[1]{\texttt{#1}}
\providecommand{\urlprefix}{URL }
\providecommand{\doi}[1]{https://doi.org/#1}

\bibitem{choi2021using}
Choi, J.W.: Using out-of-the-box frameworks for unpaired image translation and
  image segmentation for the crossmoda challenge. arXiv e-prints pp.
  arXiv--2110 (2021)

\bibitem{coelho2018mri}
Coelho, D.H., Tang, Y., Suddarth, B., Mamdani, M.: Mri surveillance of
  vestibular schwannomas without contrast enhancement: clinical and economic
  evaluation. The Laryngoscope  \textbf{128}(1),  202--209 (2018)

\bibitem{dong2021unsupervised}
Dong, H., Yu, F., Zhao, J., Dong, B., Zhang, L.: Unsupervised domain adaptation
  in semantic segmentation based on pixel alignment and self-training. arXiv
  preprint arXiv:2109.14219  (2021)

\bibitem{dorent2020scribble}
Dorent, R., Joutard, S., Shapey, J., Bisdas, S., Kitchen, N., Bradford, R.,
  Saeed, S., Modat, M., Ourselin, S., Vercauteren, T.: Scribble-based domain
  adaptation via co-segmentation. In: International Conference on Medical Image
  Computing and Computer-Assisted Intervention. pp. 479--489. Springer (2020)

\bibitem{dorent2021crossmoda}
Dorent, R., Kujawa, A., Ivory, M., Bakas, S., Rieke, N., Joutard, S., Glocker,
  B., Cardoso, J., Modat, M., Batmanghelich, K., Belkov, A., Calisto, M.B.,
  Choi, J.W., Dawant, B.M., Dong, H., Escalera, S., Fan, Y., Hansen, L.,
  Heinrich, M.P., Joshi, S., Kashtanova, V., Kim, H.G., Kondo, S., Kruse, C.N.,
  Lai-Yuen, S.K., Li, H., Liu, H., Ly, B., Oguz, I., Shin, H., Shirokikh, B.,
  Su, Z., Wang, G., Wu, J., Xu, Y., Yao, K., Zhang, L., Ourselin, S., Shapey,
  J., Vercauteren, T.: Crossmoda 2021 challenge: Benchmark of cross-modality
  domain adaptation techniques for vestibular schwannoma and cochlea
  segmentation (2022)

\bibitem{Ganin_Ustinova_Ajakan_Germain_Larochelle_Laviolette_Marchand_Lempitsky_2017}
Ganin, Y., Ustinova, E., Ajakan, H., Germain, P., Larochelle, H., Laviolette,
  F., Marchand, M., Lempitsky, V.: Domain-Adversarial Training of Neural
  Networks, p. 189–209. Advances in Computer Vision and Pattern Recognition,
  Springer International Publishing, Cham (2017),
  \url{http://link.springer.com/10.1007/978-3-319-58347-1_10}

\bibitem{Guan_Liu_2022}
Guan, H., Liu, M.: Domain adaptation for medical image analysis: A survey. IEEE
  Transactions on Biomedical Engineering  \textbf{69}(3),  1173–1185 (Mar
  2022). \doi{10.1109/TBME.2021.3117407}

\bibitem{Hoffman_Tzeng_Park_Zhu_Isola_Saenko_Efros_Darrell_2018}
Hoffman, J., Tzeng, E., Park, T., Zhu, J.Y., Isola, P., Saenko, K., Efros, A.,
  Darrell, T.: Cycada: Cycle-consistent adversarial domain adaptation. In:
  Proceedings of the 35th International Conference on Machine Learning. p.
  1989–1998. PMLR (Jul 2018),
  \url{https://proceedings.mlr.press/v80/hoffman18a.html}

\bibitem{isensee2021nnu}
Isensee, F., Jaeger, P.F., Kohl, S.A., Petersen, J., Maier-Hein, K.H.: nnu-net:
  a self-configuring method for deep learning-based biomedical image
  segmentation. Nature methods  \textbf{18}(2),  203--211 (2021)

\bibitem{Kingma_Ba_2017}
Kingma, D.P., Ba, J.: Adam: A method for stochastic optimization. arXiv
  \textbf{arXiv:1412.6980} (Jan 2017), \url{http://arxiv.org/abs/1412.6980},
  arXiv:1412.6980 [cs]

\bibitem{li2022unsupervised}
Li, H., Hu, D., Zhu, Q., Larson, K.E., Zhang, H., Oguz, I.: Unsupervised
  cross-modality domain adaptation for segmenting vestibular schwannoma and
  cochlea with data augmentation and model ensemble. In: International MICCAI
  Brainlesion Workshop. pp. 518--528. Springer (2022)

\bibitem{lastyear}
Liu, H., Fan, Y., Cui, C., Su, D., McNeil, A., Dawant, B.M.: {Unsupervised
  Domain Adaptation for Vestibular Schwannoma and Cochlea Segmentation via
  Semi-supervised Learning and Label Fusion}. In: Crimi, A., Bakas, S. (eds.)
  Brainlesion: Glioma, Multiple Sclerosis, Stroke and Traumatic Brain Injuries.
  pp. 529--539. Springer International Publishing, Cham (2022)

\bibitem{Liu_Breuel_Kautz_2017}
Liu, M.Y., Breuel, T., Kautz, J.: Unsupervised image-to-image translation
  networks. In: Advances in Neural Information Processing Systems. vol.~30.
  Curran Associates, Inc. (2017),
  \url{https://proceedings.neurips.cc/paper/2017/hash/dc6a6489640ca02b0d42dabeb8e46bb7-Abstract.html}

\bibitem{Long_Cao_Wang_Jordan_2015}
Long, M., Cao, Y., Wang, J., Jordan, M.I.: Learning transferable features with
  deep adaptation networks. arXiv  \textbf{arXiv:1502.02791} (May 2015),
  \url{http://arxiv.org/abs/1502.02791}, arXiv:1502.02791 [cs]

\bibitem{pizzini2020usefulness}
Pizzini, F.B., Sarno, A., Galazzo, I.B., Fiorino, F., Aragno, A.M., Ciceri, E.,
  Ghimenton, C., Mansueto, G.: Usefulness of high resolution t2-weighted images
  in the evaluation and surveillance of vestibular schwannomas? is gadolinium
  needed? Otology \& Neurotology  \textbf{41}(1),  e103--e110 (2020)

\bibitem{shapey2021segmentation}
Shapey, J., Kujawa, A., Dorent, R., Wang, G., Dimitriadis, A., Grishchuk, D.,
  Paddick, I., Kitchen, N., Bradford, R., Saeed, S.R., et~al.: Segmentation of
  vestibular schwannoma from mri, an open annotated dataset and baseline
  algorithm. Scientific Data  \textbf{8}(1), ~1--6 (2021)

\bibitem{shapey2019artificial}
Shapey, J., Wang, G., Dorent, R., Dimitriadis, A., Li, W., Paddick, I.,
  Kitchen, N., Bisdas, S., Saeed, S.R., Ourselin, S., et~al.: An artificial
  intelligence framework for automatic segmentation and volumetry of vestibular
  schwannomas from contrast-enhanced t1-weighted and high-resolution
  t2-weighted mri. Journal of neurosurgery  \textbf{134}(1),  171--179 (2019)

\bibitem{shin2022cosmos}
Shin, H., Kim, H., Kim, S., Jun, Y., Eo, T., Hwang, D.: Cosmos: Cross-modality
  unsupervised domain adaptation for 3d medical image segmentation based on
  target-aware domain translation and iterative self-training. arXiv preprint
  arXiv:2203.16557  (2022)

\bibitem{VS_web}
Vestibular schwannoma (acoustic neuroma) and neurofibromatosis,
  \url{https://www.nidcd.nih.gov/health/vestibular-schwannoma-acoustic-neuroma-and-neurofibromatosis}

\bibitem{wang2019automatic}
Wang, G., Shapey, J., Li, W., Dorent, R., Dimitriadis, A., Bisdas, S., Paddick,
  I., Bradford, R., Zhang, S., Ourselin, S., et~al.: Automatic segmentation of
  vestibular schwannoma from t2-weighted mri by deep spatial attention with
  hardness-weighted loss. In: International Conference on Medical Image
  Computing and Computer-Assisted Intervention. pp. 264--272. Springer (2019)

\bibitem{Yan_Ding_Li_Wang_Xu_Zuo_2017}
Yan, H., Ding, Y., Li, P., Wang, Q., Xu, Y., Zuo, W.: Mind the class weight
  bias: Weighted maximum mean discrepancy for unsupervised domain adaptation.
  In: 2017 IEEE Conference on Computer Vision and Pattern Recognition (CVPR).
  p. 945–954. IEEE, Honolulu, HI (Jul 2017). \doi{10.1109/CVPR.2017.107},
  \url{http://ieeexplore.ieee.org/document/8099590/}

\bibitem{Yu_Zhang_Dong_Hu_Dong_Zhang_2021}
Yu, F., Zhang, M., Dong, H., Hu, S., Dong, B., Zhang, L.: Dast: Unsupervised
  domain adaptation in semantic segmentation based on discriminator attention
  and self-training. Proceedings of the AAAI Conference on Artificial
  Intelligence  \textbf{35}(1212),  10754–10762 (May 2021).
  \doi{10.1609/aaai.v35i12.17285}

\bibitem{zhu2017unpaired}
Zhu, J.Y., Park, T., Isola, P., Efros, A.A.: Unpaired image-to-image
  translation using cycle-consistent adversarial networks. In: Proceedings of
  the IEEE international conference on computer vision. pp. 2223--2232 (2017)

\bibitem{zhu2022acoustic}
Zhu, Q., Li, H., Cass, N.D., Lindquist, N.R., Tawfik, K.O., Oguz, I.: Acoustic
  neuroma segmentation using ensembled convolutional neural networks. In:
  Medical Imaging 2022: Biomedical Applications in Molecular, Structural, and
  Functional Imaging. vol. 12036, pp. 228--234. SPIE (2022)

\bibitem{zou2018unsupervised}
Zou, Y., Yu, Z., Kumar, B., Wang, J.: Unsupervised domain adaptation for
  semantic segmentation via class-balanced self-training. In: Proceedings of
  the European conference on computer vision (ECCV). pp. 289--305 (2018)

\end{thebibliography}
\end{document}